\documentclass[letterpaper]{article} 
\usepackage{aaai2027}  

\nocopyright  
\usepackage[hyphens]{url}  
\usepackage{graphicx} 
\usepackage{natbib}  
\usepackage{caption} 
\usepackage{amsmath,amsfonts,amssymb}
\usepackage{booktabs}
\usepackage{array}   
\usepackage{enumitem}
\usepackage{tikz}
\usetikzlibrary{arrows.meta, positioning}

\definecolor{rsessred}{RGB}{204,0,0}

\definecolor{sessorange}{RGB}{204,102,0}

\colorlet{updres}{sessorange}

\colorlet{revtxt}{violet!85!black}
\newcommand{\trev}[1]{{\color{revtxt}#1}}
\definecolor{sesspink}{RGB}{214,51,132}
\newcommand{\pk}[1]{{\color{sesspink}#1}}
\definecolor{bsessblue}{RGB}{0,0,255}
\newcommand{\bsess}[1]{{\color{bsessblue}#1}}
\newcommand{\pending}[1]{{\color{red}\textbf{[PENDING: #1]}}}

\newif\ifstripmarkup
\stripmarkuptrue
\ifstripmarkup
  \colorlet{sessorange}{black}\colorlet{updres}{black}
  \colorlet{revtxt}{black}\colorlet{sesspink}{black}
  \colorlet{bsessblue}{black}\colorlet{rsessred}{black}
  \renewcommand{\pending}[1]{}
\fi

\graphicspath{{figures/}}

\title{Can Deep Generative Models Reproduce\\
Non-Stationary Gaussian Random Fields?} 

  \author{Daniel Kua, Yan Song}
  \affiliations{Department of Statistics, University of British Columbia}
\affiliations{
    Department of Statistics, University of British Columbia
}

\begin{document}
\maketitle

\begin{abstract}
Deep generative models (DGMs) are widely used for complex high-dimensional data and increasingly applied to spatial and spatio-temporal modeling. Their generated samples implicitly represent the learned data distribution and associated uncertainty. However, for real-world data, assessing whether DGMs have learned the underlying process is difficult because the ground truth is unknown and evaluation often relies on observations alone. We evaluate representative DGMs, flow matching (FM), DDPM, score-SDE, and VAE, on a known non-stationary Gaussian random field.
This paper provides comprehensive metrics to assess recovery of the ground-truth mean and covariance structures, with oracle samples and a stationary control as references. All four models recover the mean surface, while their covariance recovery differs across model families: DDPM and score-SDE recover the covariance structure reasonably well, FM exhibits mildly attenuated non-stationarity and slight variance under-dispersion, and VAE has difficulty recovering the covariance structure. An experiment on ERA5 temperature anomalies further demonstrates how the framework can support the validation and development of DGMs for complex real-world spatio-temporal data.
\end{abstract}


\section{Introduction}
Deep generative models (DGMs) have emerged as a powerful framework for modeling complex high-dimensional data and are increasingly used to address spatial and spatial-temporal problems in climate and environmental science \citep{corrdiff}. By generating ensembles of plausible realizations, DGMs provide an implicit representation of uncertainty. However, for real-world data, assessing whether DGMs have faithfully learned the underlying distribution, and hence whether their implied uncertainty is reliable, remains difficult because the true distribution is unknown. Evaluation therefore often relies on a limited set of summary statistics computed from observations, which a model may match while still misrepresenting the underlying dependence and variance structure \citep{hamill2001,scheuerer2015,thorarinsdottir2016}.

We address this challenge using a controlled benchmark in which training data are generated from a pre-specified Gaussian random field (GRF) or a Gaussian process, $Y(\cdot)\sim\mathcal{GP}\bigl(\mu(s),\,C(s,s')+\tau^2\delta(s,s')\bigr)$, where $\tau^2\delta(s,s')$ is the covariance contribution from pixelwise noise with $\delta(s,s')=1$ if $s=s'$ and $0$ otherwise. The GRF is designed to be non-stationary by allowing both the mean $\mu(s)$ and the dependence structure $C(s,s')$ to vary over space, yielding a challenging benchmark that reflects the heterogeneity and complexity of real-world data. With the data-generating process (DGP) known, DGMs can be evaluated systematically by assessing whether their generated samples recover key features of the target distribution, including the mean surface, non-stationary dependence structure, and variance components.

We evaluate four representative DGMs: flow matching  \citep[FM,][]{lipman2023flow}, the denoising diffusion probabilistic model \citep[DDPM,][]{ho2020ddpm}, score-based generative modeling through stochastic differential equations \citep[score-SDE,][]{song2021sde}, and the variational autoencoder  \citep[VAE,][]{kingma2014vae}. To enable recovery of spatially varying dependence, the DGM architectures use coordinate-aware inputs rather than purely convolutional, translation-equivariant networks. We also include two references: an \textbf{oracle}, consisting of independent realizations from the true GRF and representing the best achievable finite-sample performance, and a \textbf{stationary control}, defined by a GRF with stationary covariance and serving as a baseline without spatially varying dependence.

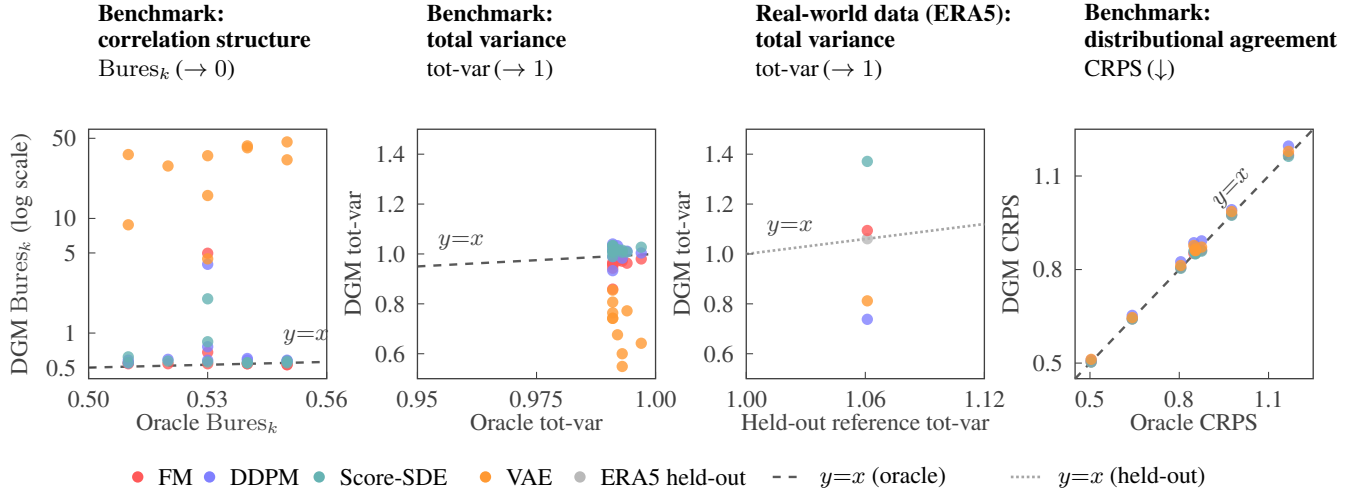
\begin{figure*}[t]
\centering
\begin{tikzpicture}[font=\small, pt/.style={font=\small\bfseries, align=left}]
\def\PW{3.15}\def\PH{3.3}\def\BY{-3.15}
\begin{scope}[xshift=0.95cm]
  \def\XS{52.5}\def\YS{1.516482}\def\LZ{0.397940}
  \node[pt, anchor=west] at (0,1.30) {Benchmark:\\correlation structure\\[1pt]\mdseries$\mathrm{Bures}_k$\,{\small ($\to 0$)}};
  \draw[black!75, line width=0.5pt] (0,\BY) rectangle (\PW,{\BY+\PH});
  \pgfmathsetmacro\ya{\BY+(log10(0.50)+\LZ)*\YS}\pgfmathsetmacro\yb{\BY+(log10(0.56)+\LZ)*\YS}
  \draw[dashed, black!65, line width=0.9pt] (0,\ya) -- (\PW,\yb);
  \node[anchor=west, font=\small, gray!50!black] at (2.45,-2.60) {$y{=}x$};
  \foreach \t in {0.50,0.53,0.56}{
    \pgfmathsetmacro\tp{(\t-0.50)*\XS}
    \draw[black!75, line width=0.4pt] (\tp,\BY) -- (\tp,{\BY+0.09});
    \node[anchor=north, font=\small, gray!50!black] at (\tp,{\BY-0.04}) {\t};
  }
  \foreach \t in {0.5,1,5,10,50}{
    \pgfmathsetmacro\tq{\BY+(log10(\t)+\LZ)*\YS}
    \draw[black!75, line width=0.4pt] (0,\tq) -- (0.09,\tq);
    \node[anchor=east, font=\small, gray!50!black] at (-0.04,\tq) {\t};
  }
  \node[anchor=north, font=\small, gray!50!black] at ({\PW/2},{\BY-0.34}) {Oracle $\mathrm{Bures}_k$};
  \node[rotate=90, anchor=south, font=\small, gray!50!black] at (-0.62,{\BY+1.65}) {DGM $\mathrm{Bures}_k$ (log scale)};
  \foreach \x/\y/\c in {%
    0.53/0.54/red!65, 0.55/0.53/red!65, 0.54/0.54/red!65, 0.52/0.54/red!65, 0.51/0.54/red!65, 0.55/0.53/red!65, 0.51/0.55/red!65, 0.54/0.54/red!65, 0.53/4.97/red!65, 0.53/0.68/red!65,
    0.53/0.58/blue!50, 0.55/0.57/blue!50, 0.54/0.58/blue!50, 0.52/0.59/blue!50, 0.51/0.58/blue!50, 0.55/0.58/blue!50, 0.51/0.55/blue!50, 0.54/0.60/blue!50, 0.53/3.98/blue!50, 0.53/0.76/blue!50,
    0.53/0.56/teal!60, 0.55/0.55/teal!60, 0.54/0.55/teal!60, 0.52/0.57/teal!60, 0.51/0.62/teal!60, 0.55/0.57/teal!60, 0.51/0.55/teal!60, 0.54/0.55/teal!60, 0.53/1.99/teal!60, 0.53/0.84/teal!60,
    0.53/35.25/orange!80, 0.55/46.49/orange!80, 0.54/41.33/orange!80, 0.52/28.68/orange!80, 0.51/36.09/orange!80, 0.55/32.47/orange!80, 0.51/8.80/orange!80, 0.54/42.94/orange!80, 0.53/4.44/orange!80, 0.53/15.85/orange!80}{
    \pgfmathsetmacro\px{(\x-0.50)*\XS}\pgfmathsetmacro\py{\BY+(log10(\y)+\LZ)*\YS}
    \fill[\c, opacity=0.8] (\px,\py) circle (0.075);
  }
\end{scope}
\begin{scope}[xshift=5.30cm]
  \def\XS{63}\def\YS{3.3}
  \node[pt, anchor=west] at (0,1.30) {Benchmark:\\ total variance\\[1pt]\mdseries tot-var\,{($\to 1$)}};
  \draw[black!75, line width=0.5pt] (0,\BY) rectangle (\PW,{\BY+\PH});
  \pgfmathsetmacro\ya{\BY+(0.95-0.5)*\YS}\pgfmathsetmacro\yb{\BY+(1.00-0.5)*\YS}
  \draw[dashed, black!65, line width=0.9pt] (0,\ya) -- (\PW,\yb);
  \node[anchor=west, font=\small, gray!50!black] at (0.15,-1.34) {$y{=}x$};
  \foreach \t in {0.95,0.975,1.00}{
    \pgfmathsetmacro\tp{(\t-0.95)*\XS}
    \draw[black!75, line width=0.4pt] (\tp,\BY) -- (\tp,{\BY+0.09});
    \node[anchor=north, font=\small, gray!50!black] at (\tp,{\BY-0.04}) {\t};
  }
  \foreach \t in {0.6,0.8,1.0,1.2,1.4}{
    \pgfmathsetmacro\tq{\BY+(\t-0.5)*\YS}
    \draw[black!75, line width=0.4pt] (0,\tq) -- (0.09,\tq);
    \node[anchor=east, font=\small, gray!50!black] at (-0.04,\tq) {\t};
  }
  \node[anchor=north, font=\small, gray!50!black] at ({\PW/2},{\BY-0.34}) {Oracle tot-var};
  \node[rotate=90, anchor=south, font=\small, gray!50!black] at (-0.62,{\BY+1.65}) {DGM tot-var};
  \foreach \x/\y/\c in {%
    0.991/0.966/red!65, 0.993/0.978/red!65, 0.992/0.972/red!65, 0.991/0.957/red!65, 0.991/0.959/red!65, 0.993/0.971/red!65, 0.997/0.980/red!65, 0.994/0.963/red!65, 0.991/0.859/red!65, 0.991/0.943/red!65,
    0.991/1.024/blue!50, 0.993/1.009/blue!50, 0.992/1.033/blue!50, 0.991/0.989/blue!50, 0.991/1.040/blue!50, 0.993/0.983/blue!50, 0.997/1.004/blue!50, 0.994/1.011/blue!50, 0.991/0.933/blue!50, 0.991/1.014/blue!50,
    0.991/1.013/teal!60, 0.993/1.015/teal!60, 0.992/1.018/teal!60, 0.991/1.006/teal!60, 0.991/1.032/teal!60, 0.993/1.007/teal!60, 0.997/1.027/teal!60, 0.994/1.008/teal!60, 0.991/0.990/teal!60, 0.991/1.034/teal!60,
    0.991/0.764/orange!80, 0.993/0.549/orange!80, 0.992/0.676/orange!80, 0.991/0.807/orange!80, 0.991/0.855/orange!80, 0.993/0.600/orange!80, 0.997/0.642/orange!80, 0.994/0.772/orange!80, 0.991/0.742/orange!80, 0.991/0.742/orange!80}{
    \pgfmathsetmacro\px{(\x-0.95)*\XS}\pgfmathsetmacro\py{\BY+(\y-0.5)*\YS}
    \fill[\c, opacity=0.8] (\px,\py) circle (0.075);
  }
\end{scope}
\begin{scope}[xshift=9.65cm]
  \def\XS{26.25}\def\YS{3.3}
  \node[pt, anchor=west] at (0,1.30) {Real-world data (ERA5):\\ total variance\\[1pt]\mdseries tot-var\,{($\to 1$)}};
  \draw[black!75, line width=0.5pt] (0,\BY) rectangle (\PW,{\BY+\PH});
  \pgfmathsetmacro\ya{\BY+(1.00-0.5)*\YS}\pgfmathsetmacro\yb{\BY+(1.12-0.5)*\YS}
  \draw[densely dotted, gray!70, line width=1.0pt] (0,\ya) -- (\PW,\yb);
  \node[anchor=west, font=\small, gray!50!black] at (0.15,-1.16) {$y{=}x$};
  \foreach \t in {1.00,1.06,1.12}{
    \pgfmathsetmacro\tp{(\t-1.00)*\XS}
    \draw[black!75, line width=0.4pt] (\tp,\BY) -- (\tp,{\BY+0.09});
    \node[anchor=north, font=\small, gray!50!black] at (\tp,{\BY-0.04}) {\t};
  }
  \foreach \t in {0.6,0.8,1.0,1.2,1.4}{
    \pgfmathsetmacro\tq{\BY+(\t-0.5)*\YS}
    \draw[black!75, line width=0.4pt] (0,\tq) -- (0.09,\tq);
    \node[anchor=east, font=\small, gray!50!black] at (-0.04,\tq) {\t};
  }
  \node[anchor=north, font=\small, gray!50!black] at ({\PW/2},{\BY-0.34}) {Held-out reference tot-var};
  \node[rotate=90, anchor=south, font=\small, gray!50!black] at (-0.62,{\BY+1.65}) {DGM tot-var};
  \foreach \x/\y/\c in {1.061/1.061/gray!55, 1.061/1.094/red!65, 1.061/0.738/blue!50,
                        1.061/1.371/teal!60, 1.061/0.812/orange!80}{
    \pgfmathsetmacro\px{(\x-1.00)*\XS}\pgfmathsetmacro\py{\BY+(\y-0.5)*\YS}
    \fill[\c, opacity=0.8] (\px,\py) circle (0.075);
  }
\end{scope}
\begin{scope}[xshift=14.00cm]
  \def\XS{3.9375}\def\YS{4.125}
  \node[pt, anchor=west] at (0,1.30) {Benchmark:\\ distributional agreement \\[1pt]\mdseries CRPS\,{\small ($\downarrow$)}};
  \draw[black!75, line width=0.5pt] (0,\BY) rectangle (\PW,{\BY+\PH});
  \draw[dashed, black!65, line width=0.9pt] (0,\BY) -- (\PW,{\BY+\PH});
  \node[rotate=46.3, anchor=south, font=\small, gray!50!black] at (2.25,-0.793) {$y{=}x$};
  \foreach \t in {0.5,0.8,1.1}{
    \pgfmathsetmacro\tp{(\t-0.45)*\XS}
    \draw[black!75, line width=0.4pt] (\tp,\BY) -- (\tp,{\BY+0.09});
    \node[anchor=north, font=\small, gray!50!black] at (\tp,{\BY-0.04}) {\t};
  }
  \foreach \t in {0.5,0.8,1.1}{
    \pgfmathsetmacro\tq{\BY+(\t-0.45)*\YS}
    \draw[black!75, line width=0.4pt] (0,\tq) -- (0.09,\tq);
    \node[anchor=east, font=\small, gray!50!black] at (-0.04,\tq) {\t};
  }
  \node[anchor=north, font=\small, gray!50!black] at ({\PW/2},{\BY-0.34}) {Oracle CRPS};
  \node[rotate=90, anchor=south, font=\small, gray!50!black] at (-0.62,{\BY+1.65}) {DGM CRPS};
  \foreach \x/\y/\c in {%
    0.854/0.854/red!65, 0.504/0.504/red!65, 0.642/0.642/red!65, 1.167/1.168/red!65, 0.805/0.806/red!65, 0.976/0.976/red!65, 0.855/0.852/red!65, 0.854/0.856/red!65, 0.849/0.852/red!65, 0.875/0.862/red!65,
    0.854/0.872/blue!50, 0.504/0.510/blue!50, 0.642/0.653/blue!50, 1.167/1.196/blue!50, 0.805/0.825/blue!50, 0.976/0.992/blue!50, 0.855/0.863/blue!50, 0.854/0.875/blue!50, 0.849/0.886/blue!50, 0.875/0.892/blue!50,
    0.854/0.852/teal!60, 0.504/0.503/teal!60, 0.642/0.641/teal!60, 1.167/1.163/teal!60, 0.805/0.804/teal!60, 0.976/0.974/teal!60, 0.855/0.854/teal!60, 0.854/0.850/teal!60, 0.849/0.859/teal!60, 0.875/0.859/teal!60,
    0.854/0.861/orange!80, 0.504/0.512/orange!80, 0.642/0.646/orange!80, 1.167/1.179/orange!80, 0.805/0.813/orange!80, 0.976/0.986/orange!80, 0.855/0.861/orange!80, 0.854/0.866/orange!80, 0.849/0.877/orange!80, 0.875/0.872/orange!80}{
    \pgfmathsetmacro\px{(\x-0.45)*\XS}\pgfmathsetmacro\py{\BY+(\y-0.45)*\YS}
    \fill[\c, opacity=0.8] (\px,\py) circle (0.075);
  }
\end{scope}
\def\LY{-4.45}
\fill[red!65]    (1.60,\LY) circle (0.075);  \node[anchor=west, font=\small] at (1.75,\LY) {FM};
\fill[blue!50]   (2.55,\LY) circle (0.075);  \node[anchor=west, font=\small] at (2.70,\LY) {DDPM};
\fill[teal!60]   (4.00,\LY) circle (0.075);  \node[anchor=west, font=\small] at (4.15,\LY) {Score-SDE};
\fill[orange!80] (6.20,\LY) circle (0.075);  \node[anchor=west, font=\small] at (6.35,\LY) {VAE};
\fill[gray!55]   (7.45,\LY) circle (0.075);  \node[anchor=west, font=\small] at (7.60,\LY) {ERA5 held-out};
\draw[dashed, black!65, line width=0.9pt] (10.00,\LY) -- (10.40,\LY);
\node[anchor=west, font=\small] at (10.50,\LY) {$y{=}x$ (oracle)};
\draw[densely dotted, gray!70, line width=1.0pt] (13.15,\LY) -- (13.55,\LY);
\node[anchor=west, font=\small] at (13.65,\LY) {$y{=}x$ (held-out)};
\end{tikzpicture}
\caption{Main DGM evaluation results under the controlled benchmark and ERA5, using selected metrics from this work and continuous ranked probability score \citep[CRPS,][]{gneiting2007} as an aggregate measure of overall distributional agreement. Each point plots a DGM value against its oracle or held-out reference counterpart, from Tabs. \ref{tab:krig}--\ref{tab:era5} and \ref{tab:dgp-sweep}--\ref{tab:dgp-sweep-direct}.}\label{fig:headline}
\end{figure*}

Our evaluation shows that all four DGMs recover the mean surface reasonably well, but differ in recovering non-stationary dependence and variance components. The first three panels of Fig.~\ref{fig:headline} summarizes these differences across all settings using selected metrics from our evaluation, with detailed analyses presented in Section~\ref{sec:experiments}. Overall, DDPM and score-SDE provide the strongest recovery. FM is close to the oracle on dominant summary metrics but shows mild under-dispersion and attenuated non-stationarity recovery under more detailed diagnostics, while VAE struggles with both aspects. In contrast, the widely used CRPS shown in the rightmost panel does not clearly separate the models, highlighting the need for a controlled benchmark and more targeted evaluation metrics. Furthermore, we complement the controlled benchmark by evaluating the DGMs on ERA5 temperature anomalies, demonstrating how the proposed framework can support practical validation and development of DGMs for complex spatial and spatio-temporal data.

To summarize, our contributions are:
\begin{enumerate}
\item \textbf{A controlled benchmark and systematic evaluation framework for DGM validation.} We develop a known-DGP benchmark based on a non-stationary GRF, enabling systematic evaluation of whether generated samples recover the target mean, non-stationary dependence, and variance structures. This reveals failure modes that can be missed by widely used summary metrics (e.g., CRPS).
\item \textbf{A comparative evaluation and analysis of representative DGM families.} We evaluate FM, DDPM, score-SDE, and VAE across multiple settings, \bsess{providing a comparative analysis of their recovery behavior and identifying distinct under-dispersion mechanisms for FM and VAE.}
\item \textbf{A real-world extension to environmental spatial data.} We adapt the evaluation framework to ERA5 temperature anomalies, illustrating its use for practical validation of DGMs when the true DGP is unavailable.
\end{enumerate}

\section{Background and Related Work}\label{sec:related}

\paragraph{DGMs for spatial and spatio-temporal applications: the validation challenge.}
DGMs are increasingly used in climate and environmental sciences, including probabilistic weather forecasting \citep{gencast}, forecast ensemble emulation \citep{seeds}, sparse data infilling \citep{sda}, and downscaling \citep{corrdiff}. Existing studies typically validate generated outputs by comparing them with observations or held-out data using summary statistics and diagnostics \citep{bulte2025,rasp2024}. However, such evaluations are inherently limited because the true data-generating distribution is unknown and summary diagnostics may fail to capture important distributional features. A controlled benchmark with a known and parametrized DGP, together with a systematic evaluation framework, is therefore needed to assess whether generated outputs align with the ground truth and recover the intended distributional features.

\paragraph{Known-process benchmarks for generative models.}
This known-process benchmarking strategy has appeared in recent work. For example, \citet{clfm,ncs,cardoso2025} validate their proposed models using tractable stationary GRFs. Although useful for controlled validation, these studies primarily use known processes to assess individual proposed methods, while stationary benchmarks do not capture the heterogeneous mean and dependence structures often present in real applications. This motivates a broader evaluation of multiple DGM families under more expressive controlled processes.

\begin{figure*}[t]
\centering
\includegraphics[width=\textwidth]{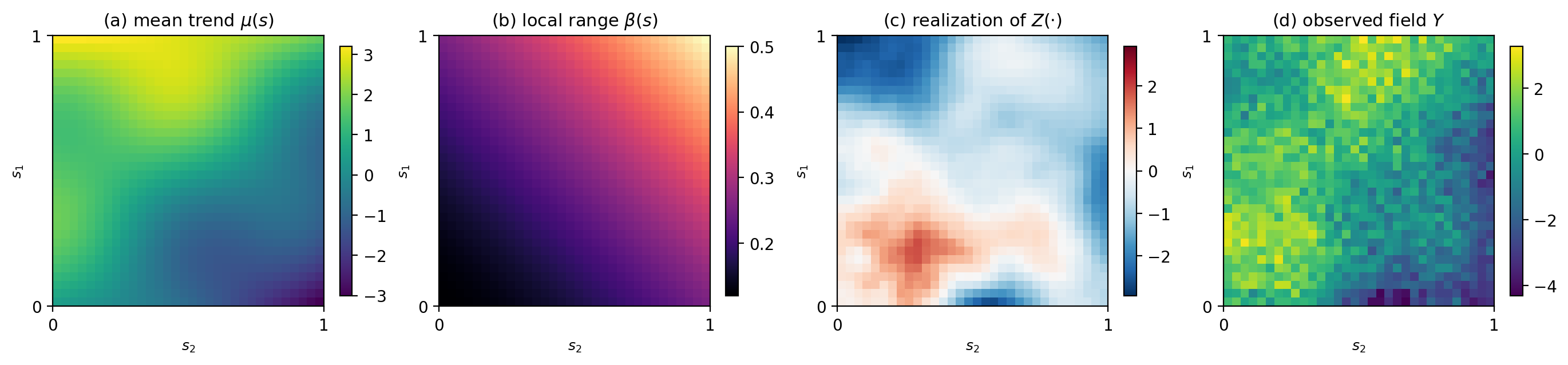}
\caption{Illustration of the DGP described in  Section~\ref{sec:dgp}. Panels (a) and (b) show the mean surface $\mu(s)$ and local range parameter $\beta(s)$, respectively. Panel (c) and (d) show one realization of the latent zero-mean spatial process $Z(\cdot)$ and the final observed field $Y(\cdot)$, respectively.}
\label{fig:dgpsummary}
\end{figure*}

\paragraph{Four generative models we test.}
We evaluate four generative model families with distinct noise-to-field generation mechanisms, providing a basis for interpreting their different recovery behaviors in Section~\ref{sec:experiments}. FM learns a deterministic velocity field and transports a noise field to data along a learned, near-straight probability path, integrating an ordinary differential equation at sampling time \citep{lipman2023flow,lipman2024guide,tong2024}. DDPM instead reverses a fixed discrete-time Gaussian noising process, learning to remove a little noise at each of many steps \citep{ho2020ddpm,karras2022edm}. Score-SDE takes the continuous-time limit, learning the score (the gradient of the log-density) of a noising SDE and integrating a reverse-time SDE with a  Langevin corrector that re-injects noise, making its sampler stochastic whereas FM uses a deterministic sampler \citep{song2021sde}. A vanilla VAE is a different family altogether: an amortized latent-variable model trained on the evidence lower bound (ELBO) that compresses each field to a low-dimensional latent and decodes it deterministically, rather than transporting a full-dimensional noise field \citep{kingma2014vae}. Together these models represent several of today's leading approaches to deep generative modelling.

\section{A controlled benchmark for evaluating DGM reproduction of GRFs}\label{sec:method}

\subsection{Testbed: A Known Non-Stationary GRF}\label{sec:dgp}

\paragraph{Ground-truth DGP.} We assume that observations on a $32{\times}32$ grid are generated from a non-stationary GRF designed to reflect key features of realistic applications. Specifically, for each grid location $s=(s_1,s_2)$, the observed value is generated as
\begin{equation}
Y(s)=\mu(s)+Z(s)+\epsilon(s),
\label{eq:model}
\end{equation}
where $\mu(s)$ is a deterministic mean trend, $Z(\cdot)\sim\mathcal{GP}\bigl(0,C(s,s')\bigr)$ is a zero-mean GRF with covariance function $C(s,s')$, and $\epsilon(s)$ is an independent nugget term representing pixel-level noise, following $\mathcal{N}(0,\tau^2)$.

\paragraph{Mean trend.} The mean trend is specified as
\begin{equation}
\mu(s)=\mathbf{c}^{\top}\mathbf{b}(s)+a\,\sin(2\pi s_1)\cos(2\pi s_2),
\label{eq:mean}
\end{equation}
where $\mathbf{b}(s)=(1,s_1,s_2,s_1^2,s_2^2,s_1 s_2)^{\top}$. The first term represents a standard polynomial trend in the spatial coordinates \citep{diggle2007}, while the second term adds a smooth non-polynomial component, making the mean surface sufficiently flexible for the benchmark \citep{cressiewikle2011}. We fix $\mathbf{c}=(0.5,1.5,-2.0,1.2,-1.5,2.0)^{\top}$ and $a=0.8$. The resulting mean surface is shown in Fig.~\ref{fig:dgpsummary}(a).

\paragraph{Covariance structure.} The GRF $Z(\cdot)$ is a stochastic process that accounts for spatial dependence. Its covariance function $C(s,s')$ describes the strength of dependence between grid points $s$ and $s'$. In this work, we adopt a non-stationary variant of the Mat\'ern covariance based on the construction of \citet{paciorek2003,paciorek2006}:
\begin{equation}
    C(s,s') = \frac{\sigma^2}{2^{\nu-1}\Gamma(\nu)}\,\frac{\beta(s)\,\beta(s')}{\bar\beta^{\,2}}\,\left(\frac{h}{\bar\beta}\right)^{\nu}K_{\nu}\left(\frac{h}{\bar\beta}\right),
    \label{eq:cov}
\end{equation}
where $h=\lVert s-s'\rVert$ is the spatial lag, $\bar\beta=\sqrt{\{\beta(s)^2+\beta(s')^2\}/2}$, $\Gamma$ is the gamma function, and $K_\nu$ is the modified Bessel function of the second kind. When $\beta(s)=\beta(s')\equiv \varphi$, this covariance reduces to the stationary Mat\'ern covariance in \eqref{eq:cov-stat}, where the variance parameter $\sigma^2$, smoothness parameter $\nu$, and range parameter $\varphi$ control the marginal variance, smoothness, and correlation decay of $Z(\cdot)$, respectively. In the non-stationary covariance \eqref{eq:cov}, $\beta(s)$ allows the dependence structure to vary across space and is referred to as the local range parameter. We set $\beta(s)=0.12+0.38\,(0.5s_1+0.5s_2)^{1.5}$; see Fig.~\ref{fig:dgpsummary}(b). Thus, $\beta(s)$ increases from $0.12$ near $(0,0)$ to $0.50$ near $(1,1)$, inducing faster correlation decay and rougher local behavior near $(0,0)$, and slower correlation decay and smoother local behavior near $(1,1)$. The resulting realizations exhibit a clear heterogeneous dependence structure; see Fig.~\ref{fig:dgpsummary}(c).


\paragraph{Main configuration and DGP robustness sweep.} The main setup fixes the covariance variance and smoothness parameters in \eqref{eq:cov}, together with the nugget variance, at $\sigma^2{=}2$, $\nu{=}1.5$, and $\tau^2{=}0.3$, respectively; see Fig.~\ref{fig:dgpsummary}(d) for one realization. We draw $50$k training realizations and $10$k validation realizations using one random seed. To check that the findings are not specific to this DGP, we repeat the full pipeline varying one parameter at a time around the base case: $\sigma^2\in\{0.5,1,2,4\}$, $\tau^2\in\{0.05,0.3,1\}$, $\nu\in\{0.5,1.5,2.5\}$, and the training-set size $n_{\text{train}}\in\{2\text{k},10\text{k},50\text{k}\}$. All models are trained under the same training budget.


\subsection{DGM Implementation Setup}\label{sec:fm}

We compare four DGM families on an equal footing: FM, DDPM, score-SDE, and VAE. Each follows a standard implementation, with key features summarized in Tab.~\ref{tab:models}.

\paragraph{FM.} FM is a continuous-time generative model that transports a noise field $x_0$ to a data realization $x_1$. We use the straight-path formulation $x_t=t\,x_1+(1-t)\,x_0$, $t\in[0,1]$, and train a neural network $v_\theta$ to predict the corresponding velocity $x_1-x_0$. After training, new realizations are generated by drawing fresh noise and integrating the learned velocity function from $t=0$ to $t=1$. We use the reference FM library without modification \citep{lipman2024guide}; each generated realization uses $100$ integration steps, corresponding to about $200$ network evaluations.

\paragraph{DDPM.} DDPM is a discrete-time diffusion model that learns to reverse a gradual Gaussian noising process. During training, a data realization is perturbed according to a cosine noise schedule \citep{nichol2021improved}, and the denoising network is trained using $v$-prediction \citep{salimans2022progressive}. Then, new realizations are generated by starting from Gaussian noise and applying $200$ deterministic DDIM reverse steps ($\eta{=}0$) \citep{song2021ddim}. We use a standard public DDPM implementation \citep{lucidrains_ddpm}, with the usual image-specific $x_0$ clipping removed because the simulated GRF values are not bounded like pixel intensities.

\paragraph{Score-SDE.} Score-SDE is a continuous-time diffusion model that learns to reverse a variance-exploding noising process. The network learns the score function of the noise-perturbed data distribution using denoising score matching. After training, new realizations are generated by starting from Gaussian noise and solving the reverse-time SDE with $100$ predictor-corrector (PC) steps \citep{song2021sde}, without image-specific clipping. We build on the official public score-SDE implementation accompanying that reference, using its SDE library verbatim.

\paragraph{VAE.} Unlike the previous three models, which transport a full-dimensional noise field, the VAE is a latent-variable generative model. During training, an encoder maps each realization to an approximate posterior distribution over the latent variable, while a decoder reconstructs the realization; training maximizes the evidence lower bound (ELBO) with the implementation's standard minibatch KL weighting (capped in the small-$n_{\text{train}}$ sweep cells; Appendix~\ref{app:repro}). After training, new realizations are generated by sampling from the latent prior and passing the sample through the decoder. We use a standard public VAE implementation \citep{pytorchvae}, with the image-specific $\tanh$ output head replaced by an unbounded output layer. The latent dimension, set to $128$, is selected as a hyperparameter, and decoding is deterministic.

\paragraph{Coordinate-aware architectures.} To enable recovery of non-stationarity, the DGM architectures are location-aware rather than purely convolutional and translation-equivariant. For FM, DDPM, and score-SDE, each network receives the spatial coordinate channels $(s_1,s_2)$ as additional inputs. For the VAE, the encoder receives these channels, while the decoder (the only network run at generation time) upsamples from a fixed-size $2{\times}2$ spatial seed computed from the latent, so absolute position is learnable by construction. Either way, the strict translation equivariance of a purely convolutional network is broken, allowing the learned model to represent location-dependent dependence structure.

\paragraph{A fair comparison.} All four DGMs are trained on the same realizations in their original scale, without standardization, and are evaluated using the same pipeline based on $n=2$k generated samples, with random seeds fixed across methods. FM, DDPM, and score-SDE use the same $4.43$M-parameter U-Net architecture. The VAE uses a comparably sized $3.9$M-parameter encoder--decoder architecture, within $12\%$ of the other models. To reduce confounding from model capacity, training budget, and sampling cost, all models are trained for the same number of optimizer steps, and the sampling budgets of FM, DDPM, and score-SDE are matched at approximately $200$ network evaluations per generated realization. The VAE generates each realization with a single decoder pass. Hyperparameters were not jointly retuned across model families; therefore, the comparison characterizes each DGM under its standard implementation rather than its best achievable calibration. More details about training and sampling can be found in Appendix~\ref{app:repro}.

\subsection{Reference Baselines}\label{sec:base}

\paragraph{Oracle.} The oracle consists of realizations drawn directly from the true GRF, representing the best achievable recovery under finite sampling. It serves as the reference level for the DGMs and as the error floor produced by finite-sample variability alone.

\paragraph{Stationary control.} The stationary control serves as a baseline without spatially varying dependence. It is obtained by fixing the local range parameter $\beta(s)$ in the true GRF to a constant value $\varphi$, while keeping all other components unchanged, resulting in the stationary covariance function
\begin{equation}
C_{\text{stat}}(s,s') = 
\frac{\sigma^2}{2^{\nu-1}\Gamma(\nu)}\,\left(\frac{h}{\varphi}\right)^{\nu}K_{\nu}\left(\frac{h}{\varphi}\right).
\label{eq:cov-stat}
\end{equation}
We set $\varphi{=}0.25$, which is close to the average value of $\beta(s)$ over the domain, approximately $0.26$. This makes the stationary control comparable to the non-stationary GRF in its overall range scale, while removing spatial variation in the dependence structure. It therefore serves as a stationary baseline for evaluating recovery of non-stationary dependence.

\subsection{Evaluation Protocol}\label{sec:eval}
We evaluate each model using complementary metrics for three aspects: mean-surface recovery ($\mathrm{MSE}_{\mu}$), non-stationary dependence recovery ($\mathrm{Bures}_k$, NS-grad, $\mathrm{corr}(\widehat\beta,\beta)$), and variance-component recovery (tot-var, $\hat\sigma^2$, $\hat\tau^2$). We describe the construction of these metrices below.

\paragraph{Mean-surface metric: $\mathrm{MSE}_{\mu}$.} We evaluate recovery of the mean surface through the mean squared error (MSE) between the generated-sample ensemble mean, $\widehat{\mu}(s)=n^{-1}\sum_{r=1}^{n}\widehat{Y}_{r}(s)$, where $\widehat{Y}_{1},\dots,\widehat{Y}_{n}$ denote the $n$ generated fields, and the ground-truth mean $\mu(s)$. Specifically,
\begin{equation*}
\mathrm{MSE}_{\mu}=\frac{1}{P}\sum_{p=1}^{P}\bigl\{\widehat{\mu}(s_p)-\mu(s_p)\bigr\}^{2},
\end{equation*}
where $P{=}1024$ is the number of grid points; smaller values indicate better recovery of the mean surface. The MSE averages over the grid and can conceal a localized failure. We therefore also inspect the un-aggregated \textbf{\emph{pixelwise mean-error map}}, $\widehat{\mu}(s_p)-\mu(s_p)$ for $p=1,\dots,P$, which shows whether the mean error carries a spatial pattern.

\paragraph{Variance-component estimates: $\widehat{\sigma}^2$ and $\widehat{\tau}^2$.} To assess whether the generated samples recover the variance components, we estimate the marginal variance $\sigma^2$ and nugget variance $\tau^2$ for samples from each generative model by fitting the ground-truth non-stationary covariance model. The smoothness parameter $\nu$ is fixed at its true value to simplify the estimation problem and reduce variability from the parameter-estimation step \citep{stein1999}. 

\begin{figure*}[t]\centering
\includegraphics[width=\textwidth]{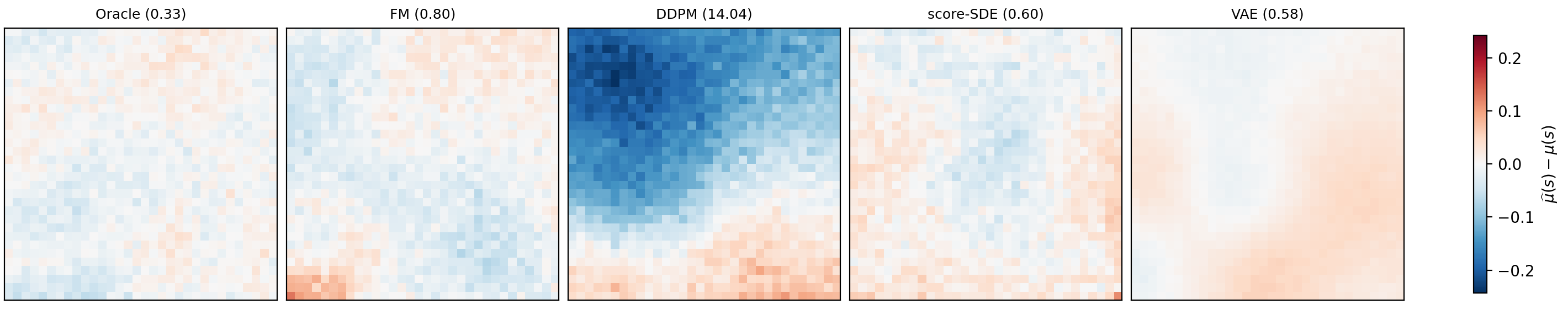}
\caption{Pixelwise mean-error maps $\widehat{\mu}(s)-\mu(s)$ for each model, with $\mathrm{MSE}_{\mu}$ $(\times 10^{-3})$ reported after each model name.}
\label{fig:meanerr}
\end{figure*}

The estimation is performed locally: we randomly subsample $500$ of the $1024$ grid points and divide the domain into $3{\times}3$ windows, each containing about $55$ of the subsampled locations. Under a local-stationarity approximation \citep{paciorek2006}, we fit a stationary Mat\'ern model within each window (Appendix~\ref{app:repro}), yielding nine local parameter estimates for each generative model. Since $\sigma^2$ and $\tau^2$ are constant in the ground-truth model, we obtain $\widehat{\sigma}^2$ and $\widehat{\tau}^2$ by averaging their local estimates across windows.

\paragraph{Non-stationarity metrics from local range estimates: NS-grad and \(\mathrm{corr}(\widehat{\beta},\beta)\).} The above procedure yields local range estimates $\widehat{\beta}(s)$ for each window, which are then used to construct two metrics for non-stationarity recovery. The first, \textbf{\emph{NS-grad}}, measures the degree of non-stationarity as the ratio between the estimated local ranges at the $(1,1)$ and $(0,0)$ corners, where the true $\beta(s)$ takes its largest and smallest values, respectively. The second, $\boldsymbol{\mathrm{corr}(\widehat{\beta},\beta)}$, is the correlation between the nine estimated local ranges and the corresponding ground-truth values of $\beta(s)$. It assesses whether the spatial variation in dependence is correctly located. 

\paragraph{Summary metrics for total variance and correlation structure: tot-var and $\mathrm{Bures}_k$.} We introduce two additional summary metrics that compare the empirical covariance matrix of generated samples, $\widehat{\Sigma}$, directly with the ground-truth covariance $\Sigma=\mathbf{C}+\tau^2\mathbf{I}$, thereby avoiding an additional covariance-parameter-estimation step.

First, the \textbf{\emph{total variance ratio}}, $\mathrm{tot\text{-}var}=\operatorname{tr}(\widehat{\Sigma})/\operatorname{tr}(\Sigma)$, checks whether the generated samples recover the overall variance level. Second, the \textbf{\emph{whitened squared Bures--Wasserstein distance}}, $\mathrm{Bures}_k$, defined as
\begin{equation*}
\mathrm{Bures}_k=\operatorname{tr}(S)+k-2\operatorname{tr}\bigl(S^{1/2}\bigr),
\end{equation*}
where $S=\Lambda_k^{-1/2}U_k^{\top}\widehat{\Sigma}\,U_k\Lambda_k^{-1/2}$ and $(U_k,\Lambda_k)$ collect the leading $k$ eigenvectors and eigenvalues of the ground-truth $\Sigma$, provides a geometric comparison of the two covariance matrices. Here $S$ is obtained by projecting $\widehat{\Sigma}$ onto the leading $k$-dimensional eigenspace of $\Sigma$ and whitening it by the corresponding eigenvalues. Then, $\mathrm{Bures}_k$ is the squared Bures--Wasserstein distance between $\mathcal{N}(0,S)$ and $\mathcal{N}(0,I_k)$ \citep{dowson1982frechet,bhatia2019bures}. Since whitening puts all $k$ directions on a common scale, this metric mainly reflects discrepancies in the dominant correlation structure and complements \(\mathrm{tot\text{-}var}\). 

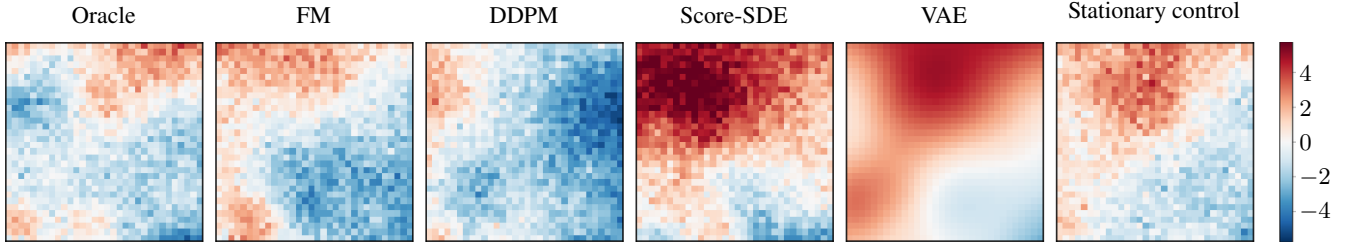
\begin{figure*}[t]
\centering

\begin{tikzpicture}[x=\textwidth, y=\textwidth, font=\small]
  \def\IW{0.96}\def\IH{0.1476}
  \node[anchor=south west, inner sep=0] at (0,0)
       {\includegraphics[width=\IW\textwidth]{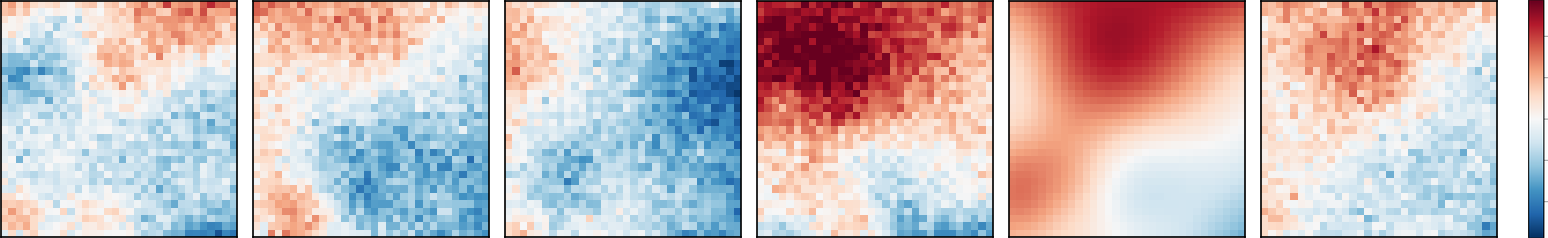}};
  \foreach \nm/\fx in {{Oracle}/0.0738, {FM}/0.2301, DDPM/0.3864,
                       Score-SDE/0.5426, VAE/0.6989, {Stationary control}/0.8552}{
    \node[anchor=south] at (\fx,{\IH+0.008}) {\nm};}
  \foreach \v/\fy in {$4$/0.8478, $2$/0.6737, $0$/0.5004, $-2$/0.3263, $-4$/0.1522}{
    \node[anchor=west, inner sep=1.5pt] at (\IW,{\fy*\IH}) {\v};}
\end{tikzpicture}
\caption{One randomly generated sample from each model under the main setting of the controlled benchmark.}
\label{fig:gridsim-col1}
\end{figure*}



\section{Experiments}\label{sec:experiments}
This section reports results for the controlled synthetic GRF benchmark (Sections~\ref{sec:meanrec}--\ref{sec:underdisp}) and a real-data experiment based on ERA5 temperature anomalies (Section~\ref{sec:era5}); 
The controlled benchmark enables direct comparison with known ground truth, while ERA5 illustrates how the evaluation framework applies when the true DGP is unavailable. For each method, we generate $n=2$k samples for evaluation; see examples in Fig.~\ref{fig:gridsim-col1}. A nearest-neighbour analysis against the training data shows no evidence of direct memorization or copying. Details of this analysis, together with computational-cost information, are provided in Appendix~\ref{sec:ExpSupple}. 

Beyond the proposed metrics, we report CRPS for comparison. Unlike our metrics, CRPS compares generated samples with observations rather than the ground truth, and provides an aggregate measure of distributional agreement rather than a targeted evaluation of dependence or variance structure. As shown in the controlled benchmark and ERA5 results below, CRPS does not clearly distinguish the model-specific differences revealed by the targeted diagnostics.

\subsection{Recovery of Mean}\label{sec:meanrec}
Figure~\ref{fig:meanerr} displays the pixelwise mean-error maps for all models, with $\mathrm{MSE}_{\mu}$ reported in each panel title. Overall, all models recover the mean surface reasonably well. FM, score-SDE, and VAE have $\mathrm{MSE}_{\mu}$ values close to the oracle finite-sample floor, whereas DDPM has a relatively larger error. The maps further reveal how mean errors vary across the domain, with most methods showing only mild spatial variation in error. FM has relatively larger errors near the $(0,0)$ corner, where the true realizations are rougher due to the shorter local range. DDPM exhibits a more coherent large-scale error pattern, with negative errors in one region and positive errors along the opposite edge. The VAE map is smoother and lower in magnitude, consistent with the smoothing effect often induced by deterministic decoding and a low-dimensional latent representation.

\subsection{Recovery of Non-Stationary Dependence}\label{sec:results}
The non-stationary dependence structure, which is rarely evaluated by commonly used pixelwise metrics, is assessed in this work using $\mathrm{Bures}_k$, NS-grad, and $\mathrm{corr}(\widehat{\beta},\beta)$. These metrics evaluate the dominant correlation structure of the generated samples, the degree of non-stationarity, and the spatial placement of local dependence variation, respectively. Tab.~\ref{tab:localrec-pooled} reports these metrics and CRPS for the main setup, and Tabs.~\ref{tab:dgp-sweep}--\ref{tab:dgp-sweep-direct} the corresponding results across every setting of the robustness sweep. Each DGM is evaluated against two references: the oracle and the stationary control.
Retraining under five random seeds demonstrates robustness across training seeds; see Appendix~\ref{sec:subsec:multiseed}.


\paragraph{Diffusion models recover the non-stationary dependence structure most faithfully, FM attenuates it, and the VAE struggles with it.} 
For $\mathrm{Bures}_k$, FM, DDPM, and score-SDE are generally close to the oracle finite-sample floor, with degradation occurring only at the smallest training size. In contrast, VAE has much larger values, indicating difficulty capturing the dominant correlation structure. 
DDPM and score-SDE also have NS-grad and $\mathrm{corr}(\widehat{\beta},\beta)$ values close to the oracle in most settings, suggesting that they recover both the degree of variation and the spatial pattern in the true local range $\beta(s)$. FM shows consistently lower NS-grad and $\mathrm{corr}(\widehat{\beta},\beta)$ values, indicating attenuated recovery of variation in $\beta(s)$, and hence of the non-stationary dependence structure compared with the diffusion models. The generally low NS-grad values and unstable, sometimes negative, $\mathrm{corr}(\widehat{\beta},\beta)$ values for the VAE indicate that it has difficulty capturing both the variation and the spatial pattern of $\beta(s)$. 

Spectral bias \citep{wang2025spectral} offers a plausible explanation: the non-stationary covariance and the nugget induce complex high-frequency behaviour in the realizations. Diffusion models may better refine this structure through multiple denoising steps, whereas the transport of FM and the smoothing decoder of VAE may struggle to capture it.

\subsection{Recovery of Variance Components}\label{sec:krig}\label{sec:underdisp}
The variance components, namely the marginal variance $\sigma^2$ of the latent spatial process $Z(\cdot)$, the nugget variance $\tau^2$, and their total, are assessed using $\widehat{\sigma}^2$, $\widehat{\tau}^2$, and tot-var, respectively. Results are summarized in Tabs.~\ref{tab:localrec-pooled}, \ref{tab:dgp-sweep} and~\ref{tab:dgp-sweep-direct}.

\paragraph{FM and the diffusion models recover both the overall variance level and its decomposition; the VAE recovers neither.} For tot-var, FM, DDPM, and score-SDE stay within a few percent of the oracle across the robustness sweep, apart from the smallest training size, where FM falls furthest below it. FM is slightly below the oracle and the diffusion models usually slightly above it, whereas VAE consistently falls short, the signature of an under-dispersed ensemble. 

The variance-component estimates provide a more detailed evaluation. For FM, DDPM, and score-SDE, both $\widehat{\sigma}^2$ and $\widehat{\tau}^2$ generally track the oracle, with FM slightly underestimating $\widehat{\sigma}^2$. In contrast, VAE collapses $\widehat{\tau}^2$ toward zero and inflates $\widehat{\sigma}^2$, suggesting that the variability of nugget $\epsilon(\cdot)$ is absorbed into the fitted process $Z(\cdot)$. Fig.~\ref{fig:gridsim-col1} supports this: the VAE sample is noticeably smoother and lacks grainy pixel-level variability. This agrees with VAE's deterministic decoder, which maps a low-dimensional latent variable to a smooth field without sampling per-pixel output noise.

\begin{table*}[t]\centering
{\small
\setlength{\tabcolsep}{4pt}
\begin{tabular}{lccc|ccc|c}
\toprule
 & \multicolumn{3}{c|}{Non-stationary dependence structure} & \multicolumn{3}{c|}{Variance components} & \multicolumn{1}{c}{Overall dist.}  \\
 \midrule
Metrics & $\mathrm{Bures}_k$ & NS-grad & corr$(\widehat\beta,\beta)$ & tot-var & $\sigma^2$ & $\tau^2$ & CRPS\,$\downarrow$ \\
\midrule
\textit{Truth} & $0$ & $2.74$ & $1$ & $1$ & $2$ & $0.3$ & $-$ \\
\midrule
Oracle & $0.532$ & $2.51_{[2.04,2.99]}$ & $0.811_{[0.762,0.864]}$ & $0.991_{[0.970,1.014]}$ &  $2.38_{[2.21,2.56]}$ & $0.288_{[0.287,0.289]}$ & $0.854$     \\
FM & $0.538$ & $1.98_{[1.66,2.35]}$ & $0.623_{[0.484,0.732]}$ & $0.966_{[0.945,0.987]}$ & $2.13_{[2.00,2.26]}$ & $0.286_{[0.285,0.287]}$ & $0.853$ \\
DDPM & $0.582$ & $2.33_{[1.90,2.84]}$ & $0.735_{[0.666,0.795]}$ & $1.024_{[1.001,1.046]}$ & $2.39_{[2.24,2.58]}$ & $0.277_{[0.276,0.278]}$ & $0.870$   \\
Score-SDE & $0.555$ & $2.95_{[2.65,3.16]}$ & $0.708_{[0.672,0.749]}$ & $1.013_{[0.991,1.036]}$ & $2.45_{[2.33,2.57]}$ & $0.289_{[0.288,0.290]}$ & $0.852$   \\
VAE & $35.25$ & $1.52_{[1.50,1.55]}$ & $0.818_{[0.805,0.832]}$ & $0.764_{[0.745,0.784]}$ & $5.23_{[5.19,5.26]}$ & $0.000_{[0.000,0.000]}$ & $0.861$    \\
\midrule
Stationary control & $5.531$ & $0.68_{[0.58,0.79]}$ & $-0.508_{[-0.619,-0.374]}$ & $0.993_{[0.970,1.016]}$ & $2.58_{[2.40,2.77]}$ & $0.287_{[0.286,0.288]}$ & $-$  \\
\bottomrule
\end{tabular}}

\caption{DGM evaluation on the controlled benchmark under the main setup. Cells are mean$_{[5\%,95\%]}$ over $B{=}n{=}2000$ field bootstraps, except for $\mathrm{Bures}_k$ and CRPS.}\label{tab:localrec-pooled}\label{tab:krig}
\end{table*}

\begin{table*}[t]
\centering
{\small
\setlength{\tabcolsep}{2pt}
\begin{tabular}{lcc|cc|cc|c}
\toprule
 & \multicolumn{2}{c|}{Marginal variance $\sigma^2(s)$} & \multicolumn{2}{c|}{Range parameter $\beta(s)$} & \multicolumn{2}{c|}{Other variance components} & \multicolumn{1}{c}{Overall dist.} \\
\cmidrule(lr){2-3}\cmidrule(lr){4-5}\cmidrule(lr){6-7}\cmidrule(lr){8-8}
Metrics & $\sigma^2$-grad & $\mathrm{corr}(\widehat{\sigma}^2,\widehat{\sigma}^2_{\mathrm{ref}})$ & NS-grad & $\mathrm{corr}(\widehat\beta,\widehat\beta_{\mathrm{ref}})$ & $\widehat{\tau}^2$ & tot-var & CRPS\,$\downarrow$ \\
\midrule
\textit{ERA5 held-out} & $49.2_{[43.9,54.6]}$ & $0.998_{[0.995,1.000]}$ & $3.97_{[3.84,4.11]}$ & $0.996_{[0.990,0.999]}$ & $0.068_{[0.055,0.081]}$ & $1.061$ & $-$ \\
\midrule
FM & $61.8_{[57.7,66.3]}$ & $0.994_{[0.989,0.997]}$ & $4.12_{[4.02,4.21]}$ & $0.993_{[0.987,0.998]}$ & $0.042_{[0.031,0.053]}$ & $1.094$ & $1.357$ \\
DDPM & $38.0_{[35.8,40.5]}$ & $0.975_{[0.968,0.981]}$ & $3.82_{[3.73,3.90]}$ & $0.949_{[0.940,0.958]}$ & $0.052_{[0.047,0.057]}$ & $0.738$ & $1.340$ \\
Score-SDE & $45.5_{[39.1,51.7]}$ & $0.996_{[0.991,0.998]}$ & $4.06_{[3.75,4.37]}$ & $0.914_{[0.892,0.936]}$ & $0.262_{[0.249,0.276]}$ & $1.371$ & $1.327$ \\
VAE & $29.8_{[26.2,32.9]}$ & $0.991_{[0.988,0.992]}$ & $3.71_{[3.64,3.79]}$ & $0.508_{[0.472,0.541]}$ & $0.000_{[0.000,0.000]}$ & $0.812$ & $1.308$ \\
\midrule
Stationary control & $1.01_{[0.96,1.06]}$ & $-0.059_{[-0.189,0.047]}$ & $2.05_{[1.56,2.50]}$ & $0.200_{[0.110,0.287]}$ & $1.742_{[1.738,1.746]}$ & $4.663$ & $-$ \\
\bottomrule
\end{tabular}}


\caption{DGM evaluation on ERA5 temperature anomalies. }
\label{tab:era5}
\end{table*}

\paragraph{Different under-dispersion mechanisms.} The under-dispersion of FM and VAE reflects different failure modes. For FM, the discrepancy appears to be mainly a small variance-scale error, for three reasons. First, its $\mathrm{Bures}_k$ value indicates that the overall correlation structure is well recovered. Second, its bootstrap interval overlaps with the oracle, suggesting that the variance deficit is a small, consistent offset rather than a sharp separation. Third, the covariance-parameter estimates place the deficit in the structured variance $\sigma^2$, rather than the nugget variance $\tau^2$. 

By contrast, the under-dispersion of VAE is not simply a global scale error. Because the missing nugget-scale variability is structural rather than a amplitude deficit, rescaling the generated samples cannot recover the pixel-level noise.



\subsection{ERA5 Temperature Anomalies}\label{sec:era5}

\paragraph{Data.} We use a $32{\times}32$ patch of ERA5 2-m temperature from WeatherBench-2 \citep{rasp2024}, covering the Northeast Pacific and western North America ($18^\circ$--$64.5^\circ$N, $195^\circ$--$241.5^\circ$E) on a $1.5^\circ$ grid, conservatively regridded from the native $0.25^\circ$ ERA5 product (Fig.~\ref{fig:gridera5}). This region is fixed a priori and exhibits spatially varying dependence from coastlines, terrain, and grid-cell shrinkage with latitude. The shrinkage affects the physical interpretation of NS-grad, but all methods share one grid, hence comparisons remain internally consistent. Each daily field is converted to a temperature anomaly by subtracting the local seasonal average estimated from the training years only; anomalies are used in Kelvin without further standardization. We split the data by year: every model is trained on 1959--2009 (${\approx}18{,}628$ fields) and evaluated against the held-out years 2015--2022 (${\approx}2{,}922$ fields). The intervening years 2010--2014 only monitor the training loss and enter no reported result.

\paragraph{Evaluation on real-world data without ground truth.} Several adjustments are needed to evaluate DGMs on ERA5. First, because the true DGP is unknown, held-out test years serve as an empirical reference while a stationary model fitted to the training data serves as the stationary control. Second, because the true covariance are unavailable for ERA5, metrics such as $\mathrm{corr}(\widehat{\beta},\beta)$ and tot-var are computed relative to estimates from the held-out reference, such as $\widehat{\beta}_{\mathrm{ref}}$ and $\widehat{\Sigma}_{\mathrm{ref}}$. Third, because ERA5 exhibits strong spatial variation in variance, we use a more general non-stationary covariance model in which both the marginal variance $\sigma^2(s)$ and the local range parameter $\beta(s)$ vary over space. In addition to NS-grad and $\mathrm{corr}(\widehat{\beta},\widehat{\beta}_{\mathrm{ref}})$, we define analogous metrics, $\sigma^2$-grad and $\mathrm{corr}(\widehat{\sigma}^2,\widehat{\sigma}^2_{\mathrm{ref}})$, to assess the degree of variation and spatial placement of the estimated marginal variance surface. The parameter estimation procedure based on local windows is identical to the benchmark, with $\nu=1.5$. For ERA5, the two windows used to compute each gradient-based metric are selected from the held-out reference as the windows with the largest and smallest estimated values, and the same pair is then fixed for all generated samples.

\paragraph{Benchmark findings largely persist on ERA5.} 
Because the training and test periods differ, temporal distribution shift may be present. The held-out estimates therefore serve as an empirical reference rather than an oracle target. We use the metrics descriptively to identify substantial departures across models. Metrics $\sigma^2$-grad, $\mathrm{corr}(\widehat{\sigma}^2,\widehat{\sigma}^2_{\mathrm{ref}})$, NS-grad, and $\mathrm{corr}(\widehat{\beta},\widehat{\beta}_{\mathrm{ref}})$ jointly assess recovery of both the magnitude and spatial pattern of non-stationary dependence. In Tab.~\ref{tab:era5}, FM, DDPM, and score-SDE are broadly comparable to the held-out reference and clearly distinct from the stationary control. The VAE exhibits weaker non-stationarity, with a substantially smaller $\sigma^2$-grad and lower $\mathrm{corr}(\widehat{\beta},\widehat{\beta}_{\mathrm{ref}})$. For the variance components, tot-var provides a more aggregated measure by combining the spatially varying variance $\sigma^2(s)$ with the nugget variance. The estimate $\widehat{\tau}^2$ specifically assesses the recovery of fine-scale variability. Score-SDE has comparatively large values for both metrics, suggesting rougher generated fields with greater overall variation. The VAE again collapses the nugget component, with $\widehat{\tau}^2$ estimated as zero.
\section{Conclusion}
We evaluated whether modern DGMs can reproduce the DGP of non-stationary spatial random fields. Using a known GRF as a controlled benchmark, we assessed DGM recovery of the mean surface, non-stationary dependence structure, and variance components against a known target. This reveals failure modes that may be missed by widely used observation-based metrics. All four models recover the mean surface reasonably well, but their covariance recovery differs substantially. DDPM and score-SDE provide the strongest recovery of non-stationary dependence and variance components. FM captures the dominant dependence structure but shows mild under-dispersion, while VAE struggles with covariance recovery. The ERA5 temperature-anomaly experiment illustrates how the framework can be adapted when the true data-generating process is unavailable. Overall, our work strengthens the validation of DGMs for spatial and spatio-temporal applications and aims to support the future development of DGMs in climate and environmental science.



\appendix
\setcounter{figure}{0}
\renewcommand{\thefigure}{A\arabic{figure}}
\setcounter{table}{0}
\renewcommand{\thetable}{A\arabic{table}}
\setcounter{dbltopnumber}{3}
\renewcommand{\dbltopfraction}{0.95}
\renewcommand{\dblfloatpagefraction}{0.5}
\renewcommand{\topfraction}{0.95}
\renewcommand{\textfraction}{0.05}
\renewcommand{\floatpagefraction}{0.5}

\section{Implementation Details}\label{app:repro}
This section provides supplementary details for Section~3.

\begin{table}[!ht]\centering

{\small
\setlength{\tabcolsep}{3pt}
\begin{tabular}{@{}>{\raggedright\arraybackslash}p{1.42cm}>{\raggedright\arraybackslash}p{2.40cm}p{1.15cm}>{\raggedright\arraybackslash}p{2.65cm}@{}}
\toprule
Model & Mechanism & Property & Implementation \\
\midrule
Oracle & Eqs.~(1)--(3) & ground truth & {---} \\
Stationary control & Eqs.~(1), (2), (4) & sta\-tionary cov. & {---} \\
\midrule
FM & transports noise to data along straight paths via a learned velocity field; samples by integrating the learned ODE & prob.-flow ODE & \texttt{flow\_\allowbreak matching}~\citep{lipman2024guide} \\
DDPM & learns to reverse a discrete-time Gaussian noising process (cosine schedule, $v$-prediction); deterministic DDIM sampling & discrete diffusion & \texttt{denoising-\allowbreak diffusion-\allowbreak pytorch}~\citep{lucidrains_ddpm} \\
Score-SDE & learns the score of a variance-exploding noising process; samples by solving the reverse-time SDE (PC) & continu\-ous SDE & \texttt{score\_\allowbreak sde\_\allowbreak pytorch}~\citep{song2021sde} \\
VAE & encoder--decoder trained on the ELBO; samples drawn from the latent prior and decoded & latent variable & \texttt{PyTorch-\allowbreak VAE}~\citep{pytorchvae} \\
\bottomrule
\end{tabular}}
\caption{Overview of the six models considered in the benchmark. The four DGMs follow their cited reference implementations.}
\label{tab:models}
\end{table}


\paragraph{DGM training and sampling details.} We provide details on DGM training and sampling in Section~3.2, with key features summarized in Tab.~\ref{tab:models}. All four DGMs are trained for $200$ epochs with batch size $256$ using the AdamW optimizer. The DGP sweep of Appendix~\ref{app:dgp-sweep} instead equalizes the budget in optimizer steps, so its cells with smaller training sets run proportionally more epochs at the same number of gradient updates. The learning rate is $2\times10^{-4}$ for FM, DDPM, and score-SDE, and $1\times10^{-3}$ for the VAE. During training, we maintain a moving average of the model weights (decay $0.999$) and use the averaged weights at sampling time. The VAE uses a $128$-dimensional latent variable with the reference implementation's KL weighting. Sampling budgets are matched at approximately $200$ network evaluations per generated realization for FM, DDPM, and score-SDE: FM integrates its ODE with $100$ midpoint steps, DDPM uses a $200$-step deterministic DDIM sampler with $\eta{=}0$, and score-SDE runs $100$ PC steps with \pk{a variance-exploding noise schedule from $\sigma_{\min}{=}0.01$ up to a data-driven $\sigma_{\max}$ (the maximum pairwise distance within a seeded $1{,}024$-field subsample of the training set, following Song's Technique~1)} and corrector signal-to-noise ratio $0.16$. The VAE generates each realization with a single decoder pass. Unless noted, every reported result comes from a single trained model per family.

\bsess{Note that we adapt the image-based samplers to unbounded Gaussian random fields. Porting the reference samplers raised two points. First, image DDPMs clamp the predicted clean image to $[-1,1]$; our fields are unbounded, hence the clamp must be removed. Without it, noise-prediction ($\varepsilon$) becomes unstable at the noisy end of the chain, where recovering the clean field divides the network's error by a very small factor: a DDPM at its lowest validation $\varepsilon$-loss still produced fields with about $180\times$ the true marginal standard deviation ($n{=}2$k draws; measured under an earlier pipeline configuration, quoted as an order of magnitude); we therefore use velocity ($v$) prediction \citep{salimans2022progressive}, which remains stable at the same budget. Second, because the law here is exactly Gaussian, we grade every generator directly against the known covariance by exact linear algebra (\texttt{evaluate\_direct.py}), compared against an oracle floor of fresh true draws.}


\paragraph{Parameter estimation details.} We provide details on the local parameter-estimation procedure proposed in Section~3.4. Within each small window, we pool all realizations into one empirical Matheron variogram \citep{matheron1963} on even bins, following scikit-gstat's conventions, and fit that package's Mat\'ern model \citep{skgstat} to it by unweighted trust-region reflective (TRF) least squares. To compute the covariance in Eq.~(4) at spatial lag $h$, we use \texttt{skgstat.models.matern(h, r, c0, s, b)} from scikit-gstat, with $b$ fitted rather than left at its default of $0$. Because the
Mat\'ern correlation used in the package's source code, i.e.\
$\tfrac{2}{\Gamma(\nu)}\big(\tfrac{\sqrt{\nu}\,h}{a}\big)^{\nu}K_\nu\!\big(\tfrac{2\sqrt{\nu}\,h}{a}\big)$,
differs from the parameterization stated in its documentation, i.e.\ $(h/a)^{\nu}K_\nu(h/a)$, where $a{=}r/2$, we use the following parameter mapping:
\begin{equation*}
\color{revtxt}
r=4\sqrt{\nu}\,\varphi,\qquad c_0=\sigma^2,\qquad s=\nu,\qquad b=\tau^2.
\end{equation*}

\section{Experimental Supplement}
\label{sec:ExpSupple}
We provide additional details for the experiments in Section~4.

\subsection{Visualization}
\label{sec:subsec:Visualization}
\begin{figure*}[!t]
\centering
\includegraphics[width=0.86\textwidth]{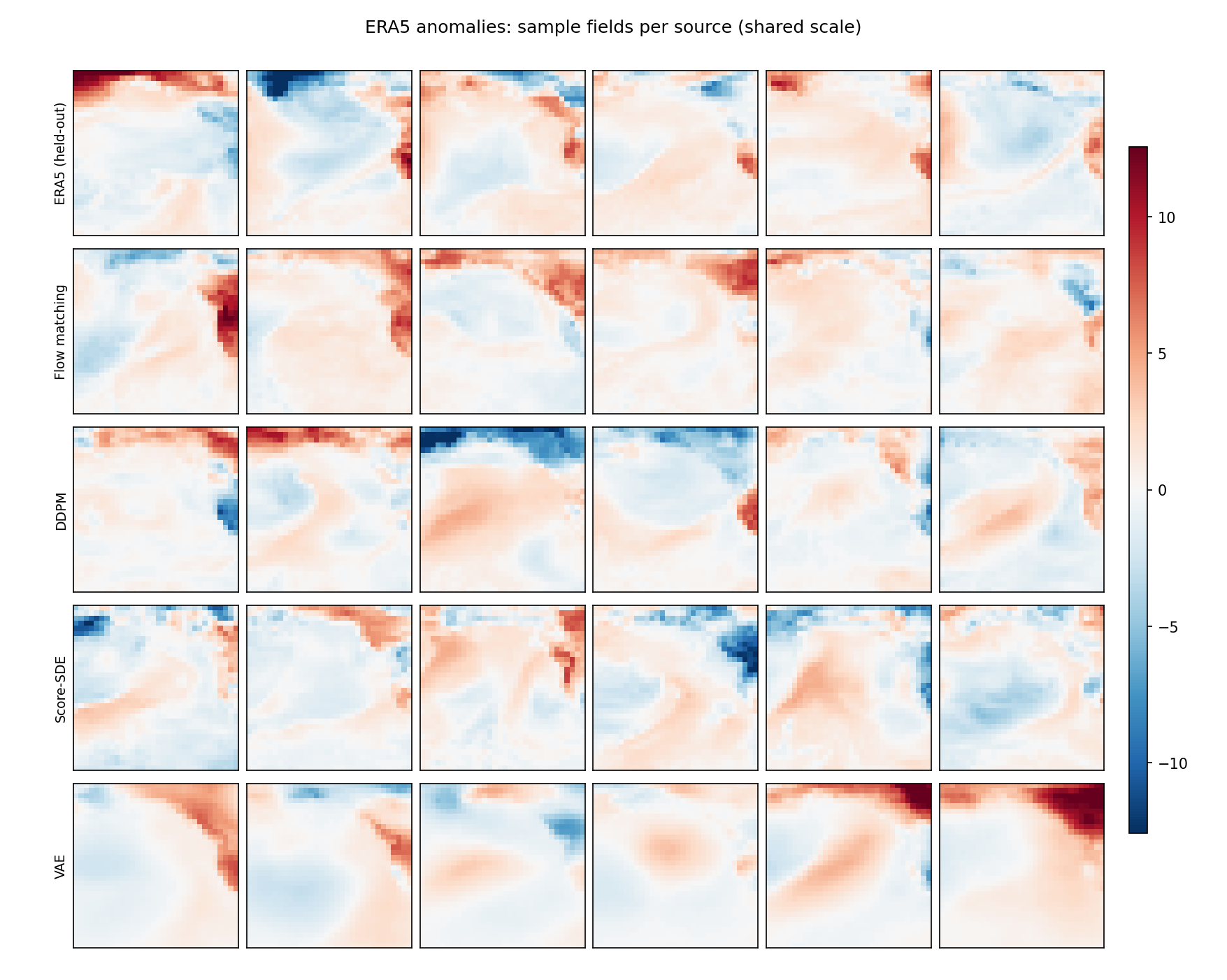}
\caption{Random samples from the ERA5 held-out reference and from the four trained DGMs.}
\label{fig:gridera5}
\end{figure*}

\begin{figure*}[!t]
\centering
\includegraphics[width=0.86\textwidth]{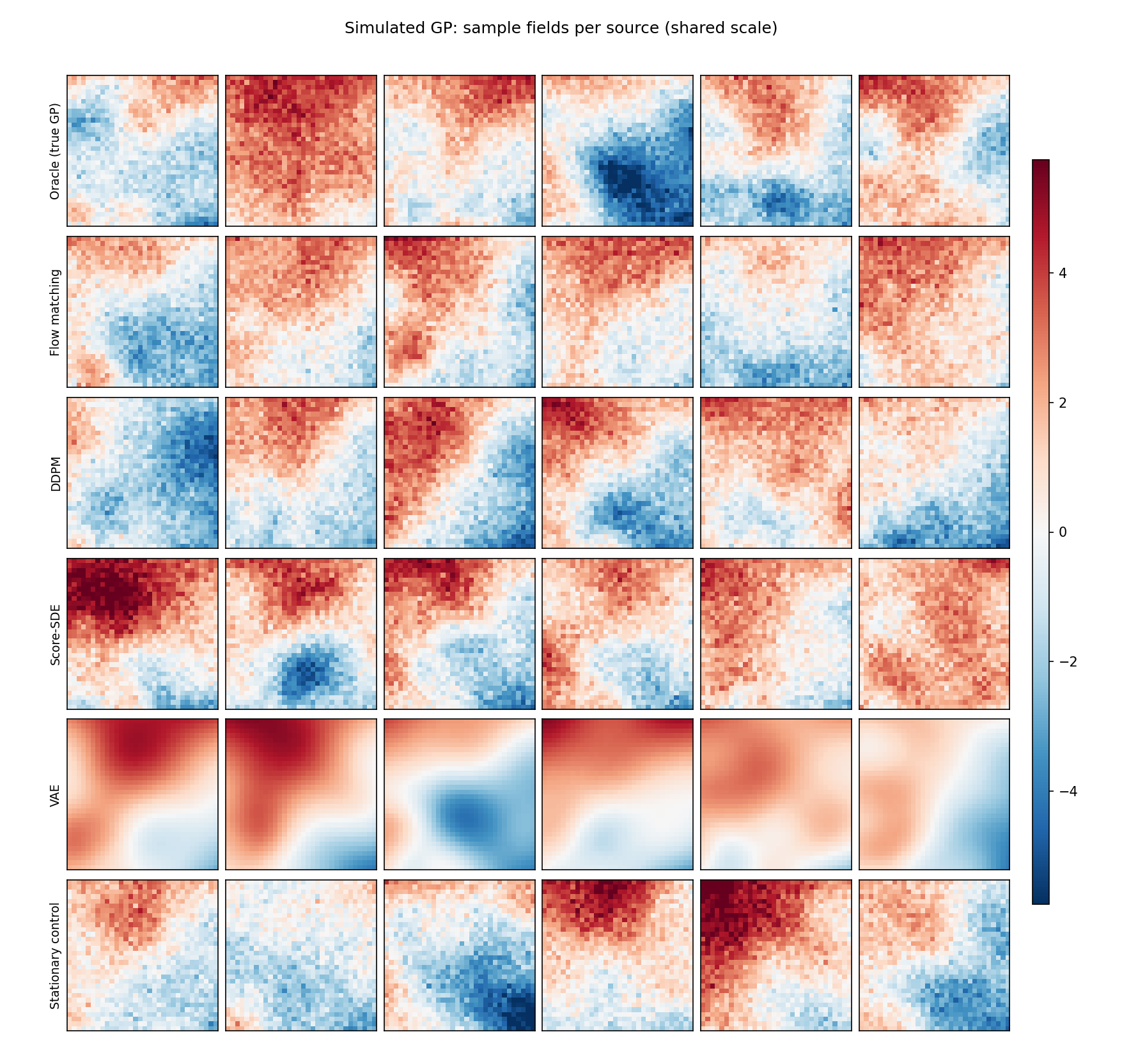}
\caption{Generated samples from the four DGMs and two reference processes under the main setting of the benchmark.}
\label{fig:gridsim}
\end{figure*}

Figures.~\ref{fig:gridera5} and \ref{fig:gridsim} display randomly generated samples for the ERA5 temperature anomalies and for the controlled benchmark.

\subsection{Common Metrics Based on Observations}
\label{sec:subsec:bench}
We introduce three observation-based metrics commonly used to assess DGMs and compare them with our ground-truth-based metrics. For each DGM, these metrics are computed using an $M$-member generated ensemble and $S$ held-out observations, with $M=50$. The number of held-out fields is $S=10$k for the controlled benchmark and $S=2{,}922$ for ERA5 (covering eight held-out years $2015$--$2022$).


\paragraph{Continuous ranked probability score \citep[CRPS,][]{gneiting2007}.} 
CRPS is a strictly proper scoring rule that assess the quality of a cumulative distribution function $F$ against an observation $y$, specifically
\begin{equation}
\mathrm{CRPS}(F,y)=\int_{-\infty}^{\infty}\big(F(z)-\mathbf{1}[z\ge y]\big)^2\,dz.
\end{equation}
For each held-out field and each grid point, we compute the unbiased fair-ensemble CRPS using the corresponding $M$-member generated ensemble \citep{ferro2014fair}, and average the resulting scores over grid points and held-out fields. It provides an aggregate measure of distributional agreement; lower values indicate better performance.

\paragraph{Mean squared error (MSE).} As with  $\mathrm{MSE}_{\mu}$ in Section 3.4, the MSE considered here evaluates the ensemble mean of the DGM-generated samples, but in a different way. Let $\bar{\hat{Y}}$ denote the ensemble mean of the generated samples and let $Y_1,\ldots,Y_S$ denote the held-out fields. We define 
\begin{equation*}
\mathrm{MSE}=\frac{1}{S P}\sum_{j=1}^{S}\big\lVert \bar{\hat Y}-Y_j\big\rVert^2.
\end{equation*}
This MSE assesses whether the DGM ensemble mean lies near the center of the held-out fields. For a fixed held-out set, the empirical minimum is achieved when $\bar{\hat Y}$ equals the held-out sample mean, and this minimum is positive because individual held-out fields vary around their mean. In the controlled benchmark, neither the true mean $\mu$ nor the oracle ensemble mean necessarily attains this empirical optimum exactly due to the finite-sample variability. Nevertheless, the oracle is expected to be close to the optimum and therefore provides a useful reference. 

\paragraph{Spread-skill ratio (SSR).} SSR evaluates ensemble reliability by comparing the ensemble spread, scaled by $\sqrt{(M{+}1)/M}$, with the square root of the MSE defined above. An ideal value is $1.0$, indicating a perfectly calibrated ensemble; values below and above $1$ indicate under- and over-dispersion, respectively. SSR is the observation-based analogue of the \trev{total-variance ratio (tot-var)}: but SSR uses held-out fields as the reference, whereas \(\mathrm{tot\text{-}var}\) uses the ground-truth variance.

\paragraph{Results.}  Tabs.~\ref{tab:bench-simreal} and \ref{tab:dgp-sweep-forecast} report results for the three observation-based metrics under the controlled benchmark and ERA5 settings. In both settings, CRPS shows limited separation among models, with all models attaining comparable values. This may reflect its role as an aggregate measure of distributional agreement.

In the controlled benchmark, DDPM exhibits mildly higher MSE values across all DGP settings, suggesting a modest deficit in mean-surface recovery. This is consistent with the finding from our $\mathrm{MSE}_{\mu}$ metric. The findings from SSR are broadly consistent with those from $\mathrm{tot\text{-}var}$. Both metrics indicate clear under-dispersion for VAE, mild under-dispersion for FM, and slight over-dispersion for score-SDE. DDPM is the main exception: SSR suggests slight under-dispersion, whereas $\mathrm{tot\text{-}var}$ does not. This discrepancy may reflect the additional finite-sample variability introduced by the held-out data in the SSR calculation.

\subsection{No Memorization or Copying}
\label{app:memorization}
For each generated field we take its nearest-neighbour distance to the $50$k training fields, summarized by the \emph{median} (robust to the skewed tail) and the \emph{minimum} (which would expose near-duplicate copies). Against fresh true-GRF draws (which cannot be copies), the median ratios are FM $1.000$, DDPM $0.998$, and score-SDE $1.010$, each with a minimum overlapping the fresh-draw minimum; memorization would pull both numbers well below the fresh-draw level, so these three produce genuinely new fields. The VAE's ratio is $0.79$, but this reflects its amplitude deficit rather than copying: an under-dispersed, smoothed field is closer to \emph{every} field, hence its whole distance distribution shifts down together, and its minimum stays at $0.81$ of its median, close to the fresh-draw shape ($0.87$), with no near-duplicate tail. The distance level is confounded for an under-dispersed model; the unchanged min-to-median shape rules out copying.

\subsection{DGP Robustness Sweep Result}
\label{app:dgp-sweep}
Tables~\ref{tab:dgp-sweep} and \ref{tab:dgp-sweep-direct} report evaluation metric values across all DGP settings described in Section~3.1, allowing us to assess the robustness of the results beyond the main setup.

\begin{table*}[!t]\centering

{\small
\setlength{\tabcolsep}{3pt}
\begin{tabular}{*{19}{>{\color{sesspink}}c}}
\toprule
\multicolumn{4}{c}{DGP configuration} &  \multicolumn{5}{c}{$\mathrm{Bures}_k$} &  \multicolumn{5}{c}{NS-grad} &  \multicolumn{5}{c}{$\mathrm{corr}(\hat\beta,\beta)$} \\
\cmidrule(lr){1-4}\cmidrule(lr){5-9}\cmidrule(lr){10-14}\cmidrule(lr){15-19}
$\sigma^2$ & $\tau^2$ & $\nu$ & $n_{\text{train}}$ & Oracle & FM & DDPM & S-SDE & VAE & Oracle & FM & DDPM & S-SDE & VAE & Oracle & FM & DDPM & S-SDE & VAE \\
\midrule
2 & 0.3 & 1.5 & 50k & 0.53 & 0.54 & 0.58 & 0.56 & 35.25 & 2.51 & 1.98 & 2.33 & 2.95 & 1.52 & 0.811 & 0.623 & 0.735 & 0.708 & 0.818 \\
\midrule
\textbf{0.5} & 0.3 & 1.5 & 50k & 0.55 & 0.53 & 0.57 & 0.55 & 46.49 & 2.17 & 1.30 & 1.70 & 2.97 & 1.00 & 0.729 & 0.155 & 0.471 & 0.610 & $-0.209$  \\
\textbf{1} & 0.3 & 1.5 & 50k & 0.54 & 0.54 & 0.58 & 0.55 & 41.33 & 2.37 & 1.63 & 2.06 & 2.95 & 1.00 & 0.816 & 0.431 & 0.660 & 0.674 & $-0.359$ \\
\textbf{4} & 0.3 & 1.5 & 50k & 0.52 & 0.54 & 0.59 & 0.57 & 28.68 & 2.44 & 2.21 & 2.40 & 2.59 & 2.43 & 0.798 & 0.700 & 0.772 & 0.746 & 0.999  \\
\midrule
2 & \textbf{0.05} & 1.5 & 50k & 0.51 & 0.54 & 0.58 & 0.62 & 36.09 & 2.64 & 2.43 & 2.58 & 2.91 & 1.50 & 0.770 & 0.720 & 0.750 & 0.746 & 0.820 \\
2 & \textbf{1} & 1.5 & 50k & 0.55 & 0.53 & 0.58 & 0.57 & 32.47 & 2.23 & 1.39 & 2.15 & 2.75 & 1.57 & 0.759 & 0.245 & 0.612 & 0.602 & 0.847 \\
\midrule
2 & 0.3 & \textbf{0.5} & 50k & 0.51 & 0.55 & 0.55 & 0.55 & 8.80 & 2.66 & 2.02 & 2.37 & 2.36 & 1.00 & 0.866 & 0.845 & 0.876 & 0.902 & $-0.464$ \\
2 & 0.3 & \textbf{2.5} & 50k & 0.54 & 0.54 & 0.60 & 0.55 & 42.94 & 1.59 & 1.03 & 1.35 & 2.01 & 1.43 & 0.367 & $-0.141$ & 0.080 & 0.285 & 0.756 \\
\midrule
2 & 0.3 & 1.5 & \textbf{2k} & 0.53 & 4.97 & 3.98 & 1.99 & 4.44 & 2.51 & 1.49 & 1.10 & 2.76 & 1.00 & 0.811 & 0.513 & 0.248 & 0.841 & $-0.050$ \\
2 & 0.3 & 1.5 & \textbf{10k} & 0.53 & 0.68 & 0.76 & 0.84 & 15.85 & 2.51 & 1.80 & 2.55 & 2.89 & 1.41 & 0.811 & 0.473 & 0.850 & 0.765 & 0.621 \\
\bottomrule
\end{tabular}}
\caption{DGM evaluation of non-stationary dependence recovery on the controlled benchmark across all DGP settings. Each row varies one parameter from the main setup in the top row, with results obtained by rerunning the full pipeline on a fresh dataset.}
\label{tab:dgp-sweep}
\end{table*}

\begin{table*}[!t]\centering

{\small
\setlength{\tabcolsep}{3pt}
\begin{tabular}{*{19}{>{\color{sesspink}}c}}
\toprule
\multicolumn{4}{c}{DGP configuration} & \multicolumn{5}{c}{tot-var} & \multicolumn{5}{c}{$\widehat\sigma^2$} & \multicolumn{5}{c}{$\widehat\tau^2$} \\
\cmidrule(lr){1-4}\cmidrule(lr){5-9}\cmidrule(lr){10-14}\cmidrule(lr){15-19}
$\sigma^2$ & $\tau^2$ & $\nu$ & $n_{\text{train}}$ & Oracle & FM & DDPM & S-SDE & VAE & Oracle & FM & DDPM & S-SDE & VAE & Oracle & FM & DDPM & S-SDE & VAE \\
\midrule
2 & 0.3 & 1.5 & 50k & 0.991 & 0.966 & 1.024 & 1.013 & 0.764 & 2.38 & 2.13 & 2.39 & 2.45 & 5.23 & 0.288 & 0.286 & 0.277 & 0.289 & 0.000 \\
\midrule
\textbf{0.5} & 0.3 & 1.5 & 50k & 0.993 & 0.978 & 1.009 & 1.015 & 0.549 & 0.70 & 0.60 & 0.56 & 0.58 & 1.21 & 0.298 & 0.297 & 0.288 & 0.299 & 0.000 \\
\textbf{1} & 0.3 & 1.5 & 50k & 0.54 & 0.54 & 0.58 & 0.55 & 41.33 & 1.24 & 1.08 & 1.19 & 1.22 & 2.97 & 0.295 & 0.293 & 0.284 & 0.296 & 0.000 \\
\textbf{4} & 0.3 & 1.5 & 50k & 0.991 & 0.957 & 0.989 & 1.006 & 0.807 & 4.60 & 4.23 & 4.67 & 4.62 & 6.00 & 0.275 & 0.273 & 0.262 & 0.274 & 0.000  \\
\midrule
2 & \textbf{0.05} & 1.5 & 50k & 0.991 & 0.959 & 1.040 & 1.032 & 0.855 & 2.33 & 2.17 & 2.42 & 2.46 & 5.17 & 0.037 & 0.036 & 0.034 & 0.037 & 0.000 \\
2 & \textbf{1} & 1.5 & 50k & 0.993 & 0.971 & 0.983 & 1.007 & 0.600 & 2.71 & 2.23 & 2.41 & 2.22 & 5.31 & 0.991 & 0.985 & 0.974 & 0.983 & 0.000 \\
\midrule
2 & 0.3 & \textbf{0.5} & 50k & 0.997 & 0.980 & 1.004 & 1.027 & 0.642 & 2.05 & 2.01 & 2.10 & 2.23 & 3.02 & 0.195 & 0.193 & 0.182 & 0.196 & 0.000  \\
2 & 0.3 & \textbf{2.5} & 50k & 0.994 & 0.963 & 1.011 & 1.008 & 0.772 & 3.50 & 2.90 & 3.03 & 3.07 & 4.75 & 0.296 & 0.294 & 0.286 & 0.297 & 0.000 \\
\midrule
2 & 0.3 & 1.5 & \textbf{2k} & 0.991 & 0.859 & 0.933 & 0.990 & 0.742 & 2.38 & 2.12 & 3.03 & 4.12 & 0.94 & 0.288 & 0.193 & 0.158 & 0.164 & 0.060  \\
2 & 0.3 & 1.5 & \textbf{10k} & 0.991 & 0.943 & 1.014 & 1.034 & 0.742 & 2.38 & 2.60 & 3.10 & 3.31 & 4.53 & 0.288 & 0.269 & 0.240 & 0.254 & 0.000 \\
\bottomrule
\end{tabular}}
\caption{DGM evaluation of variance-component recovery on the controlled benchmark across all DGP settings. Each row varies one parameter from the main setup in the top row, with results obtained by rerunning the full pipeline on a fresh dataset.}
\label{tab:dgp-sweep-direct}
\end{table*}

\begin{table*}[!t]\centering
{\small
\begin{tabular}{lcccccc}
\toprule
& \multicolumn{3}{c}{Controlled benchmark} & \multicolumn{3}{c}{ERA5 temperature anomalies} \\
\cmidrule(lr){2-4}\cmidrule(lr){5-7}
Model & CRPS\,$\downarrow$ & MSE\,$\downarrow$ & SSR\,${\to}1$ & CRPS\,$\downarrow$ & MSE\,$\downarrow$ & SSR\,${\to}1$ \\
\midrule
Oracle & 0.854 & 2.342 & 0.976 & $-$ & $-$ & $-$ \\
FM    & 0.853 & 2.328 & 0.958 & 1.357 & 8.727 & 0.943 \\
DDPM             & 0.870 & 2.436 & 0.934 & 1.340 & 8.342 & 0.828 \\
Score-SDE        & 0.852 & 2.336 & 1.046 & 1.327 & 8.856 & 1.073 \\
VAE              & 0.861 & 2.351 & 0.856 & 1.308 & 8.084 & 0.921 \\
\bottomrule
\end{tabular}}
\caption{DGM evaluation under the main setting of the controlled benchmark and ERA5 temperature anomalies, based on the observation-based metrics introduced in Appendix~\ref{sec:subsec:bench}.}
\label{tab:bench-simreal}
\end{table*}

\begin{table*}[t]\centering

{\small
\setlength{\tabcolsep}{2pt}
\begin{tabular}{*{19}{>{\color{sesspink}}c}}
\toprule
\multicolumn{4}{c}{DGP configuration} & \multicolumn{5}{c}{CRPS $\downarrow$} & \multicolumn{5}{c}{MSE $\downarrow$} & \multicolumn{5}{c}{SSR $\to 1$} \\
\cmidrule(lr){1-4}\cmidrule(lr){5-9}\cmidrule(lr){10-14}\cmidrule(lr){15-19}
$\sigma^2$ & $\tau^2$ & $\nu$ & $n_{\text{train}}$ & Oracle & FM & DDPM & S-SDE & VAE & Oracle & FM & DDPM & S-SDE & VAE & Oracle & FM & DDPM & S-SDE & VAE \\
\midrule
2 & 0.3 & 1.5 & 50k & 0.854 & 0.854 & 0.872 & 0.852 & 0.861 & 2.343 & 2.333 & 2.444 & 2.336 & 2.352 & 0.976 & 0.957 & 0.933 & 1.045 & 0.856  \\
\midrule
\textbf{0.5} & 0.3 & 1.5 & 50k & 0.504 & 0.504 & 0.510 & 0.503 & 0.512 & 0.815 & 0.812 & 0.835 & 0.814 & 0.813 & 0.981 & 0.973 & 0.949 & 1.029 & 0.781  \\
\textbf{1} & 0.3 & 1.5 & 50k & 0.642 & 0.642 & 0.653 & 0.641 & 0.646 & 1.325 & 1.319 & 1.369 & 1.323 & 1.315 & 0.978 & 0.966 & 0.943 & 1.039 & 0.843  \\
\textbf{4} & 0.3 & 1.5 & 50k & 1.167 & 1.168 & 1.196 & 1.163 & 1.179 & 4.379 & 4.358 & 4.586 & 4.354 & 4.398 & 0.975 & 0.953 & 0.909 & 1.046 & 0.874  \\
\midrule
2 & \textbf{0.05} & 1.5 & 50k & 0.805 & 0.806 & 0.825 & 0.804 & 0.813 & 2.087 & 2.076 & 2.194 & 2.082 & 2.102 & 0.975 & 0.951 & 0.926 & 1.058 & 0.880  \\
2 & \textbf{1} & 1.5 & 50k & 0.976 & 0.976 & 0.992 & 0.974 & 0.986 & 3.057 & 3.046 & 3.148 & 3.051 & 3.031 & 0.980 & 0.967 & 0.932 & 1.030 & 0.791  \\
\midrule
2 & 0.3 & \textbf{0.5} & 50k & 0.855 & 0.852 & 0.863 & 0.854 & 0.861 & 2.345 & 2.326 & 2.390 & 2.344 & 2.323 & 0.994 & 0.983 & 0.960 & 1.037 & 0.817  \\
2 & 0.3 & \textbf{2.5} & 50k & 0.854 & 0.856 & 0.875 & 0.850 & 0.866 & 2.341 & 2.342 & 2.461 & 2.327 & 2.366 & 0.967 & 0.951 & 0.914 & 1.045 & 0.854  \\
\midrule
2 & 0.3 & 1.5 & \textbf{2k} & 0.849 & 0.852 & 0.886 & 0.859 & 0.877 & 2.301 & 2.288 & 2.477 & 2.380 & 2.356 & 0.998 & 0.907 & 0.880 & 1.021 & 0.769  \\
2 & 0.3 & 1.5 & \textbf{10k} & 0.875 & 0.862 & 0.892 & 0.859 & 0.872 & 2.446 & 2.374 & 2.544 & 2.370 & 2.386 & 1.048 & 0.938 & 0.909 & 1.042 & 0.847  \\
\bottomrule
\end{tabular}}
\caption{DGM evaluation under all DGP settings of the controlled benchmark, based on the observation-based metrics introduced in Appendix~\ref{sec:subsec:bench}. Each row varies one parameter from the main setup in the top row, with results obtained by rerunning the full pipeline on a fresh dataset.}
\label{tab:dgp-sweep-forecast}
\end{table*}

\subsection{Multi-Seed Robustness}
\label{sec:subsec:multiseed}
The bootstrap CIs reported elsewhere condition on a fixed trained checkpoint. To assess training-run variability, we retrain each model family under five random seeds (\texttt{run\_multiseed.sh}) and report the across-seed mean $\pm$SD in Tab.~\ref{tab:multiseed}. Metrics based on local parameter estimation vary more across seeds than suggested by the single-checkpoint bootstrap, especially for NS-grad. However, the main conclusions persist across retrainings: FM's $\hat\sigma^2$ remains below the oracle's, DDPM's and VAE's remain above it, and VAE consistently shows nugget collapse and an attenuated gradient. A less stable case is score-SDE's $\hat\sigma^2$, which is slightly above the oracle for the single checkpoint but slightly below it when averaged across seeds. Since this shift is comparable to the across-seed SD, we do not claim a consistent direction for score-SDE.

\begin{table*}[!t]\centering
{\small
\begin{tabular}{lccccc}
\toprule
Ensemble & $\hat\sigma^2$ & $\hat\tau^2$ & tot-var & NS-grad & corr$(\widehat\beta,\beta)$ \\
\midrule
\textit{Truth} & $2$ & $0.3$ & $1$ & $2.74$ & $1$ \\
Oracle & $2.38{\pm}0.00$ & $0.288{\pm}0.000$ & $0.991{\pm}0.000$ & $2.51{\pm}0.00$ & $0.811{\pm}0.000$ \\
FM & $2.29{\pm}0.14$ & $0.287{\pm}0.000$ & $0.974{\pm}0.018$ & $2.58{\pm}0.39$ & $0.711{\pm}0.051$ \\
DDPM & $2.58{\pm}0.16$ & $0.276{\pm}0.001$ & $1.033{\pm}0.013$ & $2.79{\pm}0.30$ & $0.707{\pm}0.020$ \\
Score-SDE & $2.29{\pm}0.13$ & $0.290{\pm}0.001$ & $1.010{\pm}0.007$ & $2.38{\pm}0.36$ & $0.716{\pm}0.051$ \\
VAE & $5.19{\pm}0.03$ & $0.000{\pm}0.000$ & $0.776{\pm}0.013$ & $1.52{\pm}0.02$ & $0.819{\pm}0.008$ \\
\bottomrule
\end{tabular}}
\caption{Across-seed recovery under the main setting of the controlled benchmark. Cells report mean$\pm$SD over five training seeds per family, with $n{=}2$k generated fields and $B{=}300$ bootstrap samples per seed. Evaluation draws are fixed, so the oracle has zero across-seed variation.}
\label{tab:multiseed}
\end{table*}


\subsection{Computational Cost}
\label{sec:subsec:CompCost}
Experiments target a single NVIDIA V100 GPU with 16\,GB of memory. A representative recovery run, including data generation, FM training for $200$ epochs on $50$k fields, sampling, and bootstrap re-fits, takes approximately $66$ minutes, with runtime dominated by training. Bootstrap confidence intervals are computed by resampling $n=2$k generated fields $B=2$k times; the repeated covariance-parameter refits add CPU-bound runtime. The full pipeline (with all four generators) fits within a single 18-hour job, and the DGP robustness sweep is run as a 10-cell GPU array. The across-seed study in Tab.~\ref{tab:multiseed} adds approximately \(7\) GPU-hours per model family, except for VAE, which requires about \(2\) GPU-hours.

\section*{Reproducibility Statement}
All code to reproduce every experiment is released as a single package. Core requirements are \texttt{torch}, \texttt{numpy}, \texttt{scipy}, \texttt{torchdiffeq}, and \texttt{scikit-gstat} (for estimating covariance parameters), plus \texttt{xarray}/\texttt{zarr}/\texttt{gcsfs} for the ERA5 setting. Every model is backed by its reference implementation, cloned automatically at setup: \texttt{flow\_matching} (FM), \texttt{denoising-diffusion-pytorch} (DDPM), \texttt{score\_sde\_pytorch} (score-SDE), and \texttt{PyTorch-VAE} (\texttt{VanillaVAE}); the thin adaptation layers (coordinate conditioning, removed clamps, unbounded heads) are documented in-line. \textbf{Data.} The benchmark data are drawn from the GRF introduced in Section~3.1 for all settings, with every draw seeded for reproducibility.
ERA5 is retrieved without credentials from public WeatherBench-2 \citep{rasp2024}. \textbf{Runs.} When rerun end to end from one seed, the released pipeline regenerates the recovery tables, baseline comparisons, fix and DGP sweeps, and the ERA5 evaluation (including CRPS, MSE, and SSR). Compute details are in Appendix~\ref{sec:subsec:CompCost}.

\end{document}